\documentclass{article}

\usepackage{arxiv}
\usepackage[utf8]{inputenc}
\usepackage[T1]{fontenc}
\usepackage[hidelinks]{hyperref}
\usepackage{url}

\usepackage{subcaption}
\usepackage{amsfonts}
\usepackage{amsmath}
\usepackage{nicefrac}
\usepackage{microtype}
\usepackage{cleveref}
\usepackage{algorithm}
\usepackage{algorithmic}
\usepackage{lipsum}
\usepackage{graphicx}
\usepackage{natbib}
\usepackage{doi}
\usepackage{graphicx}
\usepackage{booktabs}
\usepackage{multirow}
\usepackage{threeparttable}
\usepackage[table]{xcolor}
\usepackage{colortbl}
\usepackage{xcolor}

\definecolor{lightgray}{gray}{0.92}

\title{EvolVE: Evolutionary Search for LLM-based Verilog Generation and Optimization}

\author{
    Wei-Po Hsin\thanks{Equal contribution} \\
	Department of Electrical Engineering \\
    National Taiwan University \\
	\texttt{b10901053@ntu.edu.tw} \\
	\And
    Ren-Hao Deng\footnotemark[1] \\
	Department of Computer Science and \\
    Information Engineering \\
    National Taiwan University \\
	\texttt{r13922212@csie.ntu.edu.tw}
    \And
    Yao-Ting, Hsieh\footnotemark[1] \\
	Institute of Information Science \\
    Academia Sinica, Taiwan \\
	\texttt{elton731888@gmail.com}
    \And
    En-Ming Huang \\
	Department of Computer Science and \\
    Information Engineering \\
    National Taiwan University \\
	\texttt{r13922078@csie.ntu.edu.tw}
    \And
    Shih-Hao Hung\thanks{Corresponding author} \\
	Department of Computer Science and \\
    Information Engineering \\
    National Taiwan University \\
	\texttt{hungsh@csie.ntu.edu.tw}
}

\renewcommand{\shorttitle}{EvolVE: Evolutionary Search for LLM-based Verilog Generation and Optimization}

\begin{document}
\maketitle
\begin{abstract} 
Verilog's design cycle is inherently labor-intensive and necessitates extensive domain expertise. Although Large Language Models (LLMs) offer a promising pathway toward automation, their limited training data and intrinsic sequential reasoning fail to capture the strict formal logic and concurrency inherent in hardware systems. To overcome these barriers, we present \textbf{EvolVE}, the first framework to analyze multiple evolution strategies on chip design tasks, revealing that Monte Carlo Tree Search (MCTS) excels at maximizing functional correctness, while Idea-Guided Refinement (IGR) proves superior for optimization. We further leverage Structured Testbench Generation (STG) to accelerate the evolutionary process. To address the lack of complex optimization benchmarks, we introduce \textbf{IC-RTL}, targeting industry-scale problems derived from the National Integrated Circuit Contest. Evaluations establish EvolVE as the new state-of-the-art, achieving 98.1\% on VerilogEval v2 and 92\% on RTLLM v2. Furthermore, on the industry-scale IC-RTL suite, our framework surpasses reference implementations authored by contest participants, reducing the Power, Performance, Area (PPA) product by up to 66\% in Huffman Coding and 17\% in the geometric mean across all problems. The source code of the IC-RTL benchmark is available at \url{https://github.com/weiber2002/ICRTL}. 
\setcounter{footnote}{0}
\footnote{The EvolVE codebase, Mod-VerilogEval v2, and evaluation scripts will be released shortly.}

\end{abstract}


\section{Introduction}

As digital hardware grows increasingly complex, the industry faces a critical need for faster development cycles without sacrificing circuit performance. While Hardware Description Languages (HDLs) like Verilog are the standard for providing fine-grained control over synchronous circuits, relying solely on manual RTL coding has become a significant bottleneck as the process is inherently labor-intensive and prone to delays. To overcome these limitations, a promising new paradigm has emerged: leveraging Large Language Models (LLMs) to automate and accelerate hardware design and verification.

Fundamentally, LLMs are powerful probabilistic models trained on vast text and code corpora. Their fundamental nature is to learn complex statistical patterns to predict the next token in a sequence, a capability that enables logical, step-by-step reasoning. Recent advancements \citep{alphaevolve, dolphin} have further enhanced this potential through advanced reasoning techniques and evolutionary frameworks, enabling agents to iteratively refine their outputs and solve intricate logical problems. However, this proficiency in sequential reasoning lacks alignment with the requirements of hardware design. HDLs like Verilog describe highly concurrent systems, where thousands of operations execute in parallel, synchronized by a clock. This strict and concurrent paradigm poses a significant hurdle for probabilistic models trained on sequential logic. Consequently, direct application of LLMs to HDL generation often results in functionally incorrect or non-synthesizable Verilog. This performance gap underscores the insufficiency of the base model alone, with the true promise for IC design residing in novel frameworks that enable LLMs to reason about concurrency, formally explore the design space, and iteratively optimize results.

Building on this realization, recent works have emerged to address the limitations of standalone LLMs in hardware design. These efforts can be categorized into two main streams. The first focuses on RTL specialization, creating expert models such as RTLCoder \citep{rtlcoder}, ScaleRTL \citep{scalertl}, and CodeV-R1 \citep{codev} through supervised fine-tuning or reinforcement learning on Verilog-specific datasets, including Pyranet \citep{pyranet} and MGVerilog \citep{mgverilog}. The second, more recent stream, employs multi-agent systems to decompose the complex design process. These frameworks orchestrate LLMs for various roles, such as leveraging intermediate representations \citep{symrtlo, rtlrewriter}, exploring diverse design candidates \citep{mage}, or managing hierarchical pipelines of specialized agents \citep{veriopt, rtlsquad}. Some even optimize the workflow itself \citep{vflow}.

Despite these advancements, accurate assessment remains a bottleneck. To evaluate methodology efficacy, the community established standardized benchmarks, such as RTLLM v2 \citep{rtllm}, for natural language specification-to-RTL generation. Concurrently, the VerilogEval v2 effort \citep{verilogeval2} extended the original benchmark from code-completion to specification-to-RTL tasks. However, our analysis reveals inconsistencies in the golden models and problem specifications of VerilogEval v2 that degrade evaluation precision. Moreover, we find that RTLLM v2 lacks the complexity of industrial designs, failing to challenge models on complex optimization techniques. To overcome these validation gaps, we introduce \textbf{Mod-VerilogEval v2}, a rectified standard benchmark detailed in Appendix~\ref{sec:modified-verilogeval2}. Additionally, we present \textbf{IC-RTL}, a high-complexity suite derived from the Taiwan National IC Design Contest \citep{TaiwanICContest} and expert-crafted custom designs.

To address the challenges of automated Verilog generation, given the scarcity of domain-specific training data and the inherent reasoning limitations of fixed-capacity models, we propose a flexible framework without fine-tuning that leverages evolutionary algorithms to unify design exploration and debugging, effectively mirroring the iterative refinement process natural to hardware design. To mitigate the computational latency typically associated with evolutionary frameworks, we introduce \textbf{Structured Testbench Generation (STG)}.

By automatically categorizing essential signals to generate rigorous test vectors, STG provides high-coverage test cases and fine-grained feedback that goes beyond simple binary results. This mechanism effectively transfers the heavy computational cost of iterative debugging from expensive LLM reasoning to efficient, free EDA simulation tools, thereby simultaneously accelerating the evolutionary process and significantly increasing the framework's capabilities. The main contributions of this paper are summarized as follows:

\begin{itemize}
    \item \textbf{EvolVE Framework:} We introduce an evolutionary framework that utilizes two distinct search strategies: Idea-Guided Refinement (IGR) and Monte Carlo Tree Search (MCTS). By integrating these search mechanisms, our framework enables LLMs to reason about concurrency and iteratively explore the design space without relying on massive domain-specific datasets or requiring prohibitively large model scales. This approach facilitates targeted self-correction, enabling DeepSeek-R1-FP4 to achieve 92\% on RTLLM v2 and 98.1\% on VerilogEval v2, while boosting the performance of smaller models like Siliconmind-7B to 92.3\% on the latter.
    
    \item \textbf{IC-RTL and Verilog Optimization:} To address the scarcity of rigorous evaluation of PPA, we further validate our framework on our novel IC-RTL benchmark suite. In this high-complexity environment, our framework outperforms reference implementations authored by contest participants, reducing the PPA product by up to 66\% in Q5 and 17\% in the geometric mean across all problems.
    
    \item \textbf{Structured Testbench Generation (STG):} We develop STG, a detachable and automated high-coverage verification engine designed to accelerate the evolutionary cycle. By automatically categorizing signals and providing fine-grained feedback, STG transfers the cost of iterative debugging to free EDA tools. 
\end{itemize}

\section{Background and Motivation}

\subsection{Hardware Description Languages}
HDLs bridge the gap between behavioral intent and physical implementation. Currently, the industry heavily relies on Verilog for chip design, as it provides the granularity needed for fine-grained optimization, ensuring that synthesized circuits meet strict Performance, Power, and Area (PPA) targets. Furthermore, the ubiquity of Verilog has created a vast corpus of existing designs that far exceeds that of newer languages such as Chisel and SystemC for High-Level Synthesis. This comparative abundance of data makes Verilog the optimal candidate for training customized LLMs for HDL generation and for leveraging LLMs to conduct automated PPA optimizations.

\subsection{LLMs for Code Generation}
Leveraging LLMs as automated coding agents has evolved into a pivotal branch of research. Initial tools like Microsoft Copilot and Aider \citep{aider} demonstrated strong proficiency in sequential languages such as Python and C++. However, while these single-turn agents perform well on standard benchmarks, they often struggle with the complexity of project-level engineering and long-horizon problem solving. To address these limitations, research has shifted toward sophisticated multi-agent frameworks. Systems such as MetaGPT \citep{metagpt} and AgileCoder \citep{agilecoder} mimic human agile methodologies, assigning diverse roles (e.g., architect, engineer, QA) to distinct agents to enable collaborative software development.

Recently, the emergence of evolutionary coding agents from the domain of automated scientific discovery has been observed. Frameworks such as AlphaEvolve \citep{alphaevolve} and Dolphin \citep{dolphin} move beyond static prompting by leveraging closed-loop optimization. In these systems, agents iteratively conduct reasoning, generate implementation code, and utilize execution feedback to refine their solutions. For instance, AlphaEvolve employs an evolutionary pipeline to discover novel algorithms, while Dolphin automates the research loop of idea generation, practice, and feedback. This shift from generation to iterative evolution provides the foundational methodology for solving the strict constraints of hardware design.

\subsection{LLMs for Verilog Generation}
While LLMs excel at step-by-step reasoning, they struggle to model the massive parallelism and cycle-accurate synchronization required by Verilog. To bridge this gap, recent efforts utilize two primary strategies: domain adaptation via fine-tuning and structural guidance via agentic frameworks.

\paragraph{Domain-Specific Fine-Tuning.}
This approach embeds hardware domain knowledge directly into the model weights. RTLCoder \citep{rtlcoder} established a baseline by generating Verilog data pairs with a GPT-assisted feedback quality scheme, achieving performance comparable to GPT-4. ScaleRTL \citep{scalertl} advanced this by leveraging Chain-of-Thought \citep{chainofthought} reasoning traces to fine-tune models for complex logic.
More recent efforts incorporate Reinforcement Learning (RL). VeriReason \citep{verireason} utilizes Group Relative Policy Optimization \citep{grpo} to enhance reasoning, while CodeV-R1 \citep{codev} employs a testbench generator for round-trip data synthesis. However, despite these specialized architectures, fine-tuned models often fail to consistently outperform top-tier proprietary models due to limitations in model scale and the persistent scarcity of high-quality, open-source Verilog training data compared to the vast corpora available for software languages.

\paragraph{Agentic and Search-Based Frameworks.}
In parallel with fine-tuning, a second stream guides general-purpose LLMs by structuring the problem-solving workflow. This can be categorized into three methodologies:

\begin{itemize}
    \item \textbf{Optimization and Debugging:} Early frameworks focused on local code rectification. RTLRewriter \citep{rtlrewriter} leverages Cost-aware Monte Carlo Tree Search (C-MCTS) to rewrite Verilog, while SymRTLO \citep{symrtlo} employs a neuro-symbolic framework using Abstract Syntax Trees (ASTs) to enforce PPA constraints.
    
    \item \textbf{Collaborative Multi-Agent Systems}: Inspired by human design teams, these frameworks decompose complex specifications into manageable sub-tasks. MAGE \citep{mage} utilizes specialized agents with a checkpoint mechanism for state tracking. RTLSquad \citep{rtlsquad} divides the workflow into exploration, implementation, and evaluation squads to generate interpretable decision paths, while VeriOpt \citep{veriopt} integrates PPA constraints via role-based prompting. VerilogCoder \citep{verilogcoder} advances this approach by utilizing a Task and Circuit Relation Graph (TCRG) to decompose module descriptions into sub-tasks, coupled with an AST-based waveform tracing tool that enables agents to autonomously debug and fix functional errors.

    \item \textbf{Workflow and Evolutionary Search}: Recent advancements treat generation as a global search problem. VFlow \citep{vflow} optimizes the agentic workflow itself, using cooperative evolution (CEPE-MCTS) to discover the optimal sequence of LLM calls. Similarly, REvolution \citep{revolution} applies a dual-population algorithm and evolutionary prompt selection to conduct both generation and optimization of the Verilog design.
\end{itemize}

\subsection{The Complexity Gap of Benchmarking PPA} \label{sec:benchmark_ppa_gap}
Currently, optimization efforts primarily target the RTLLM v2 benchmark, such as those seen in REvolution \citep{revolution}. However, these are often restricted to small-scale modules that can be trivially optimized by standard synthesis tools without deep domain knowledge. To make automated design applicable for both academia and industry, the need for complex, large-scale benchmarks is urgent. Such benchmarks must support both commercial EDA tools for precise assessment and open-source tools for broad accessibility, while providing initial reference designs to rigorously evaluate the true hardware optimization capabilities of different frameworks.

\subsection{Current Limitations and Our Approach}
While fine-tuned models demonstrate impressive capabilities, they remain constrained by limited model capacity and the scarcity of high-quality, open-source Verilog data. Reinforcing these concerns, a recent survey \citep{llmverilogsurvey} highlights specific hurdles: existing benchmarks are small-scale ($<$100 samples) and focus on basic modules, neglecting system-level complexities. Furthermore, current metrics often rely solely on functional pass rates (functional-pass@k), ignoring critical dimensions like PPA. Finally, reliance on single-turn generation restricts architectural exploration and limits the capacity to recover from logic failures.

To overcome the dual bottlenecks of data scarcity and limited model scale, we shift from a learning-centric to a search-based paradigm. Since high-quality Verilog data is proprietary and finite, and scaling model parameters yields diminishing returns without corresponding data growth, we instead leverage evolutionary search to utilize test-time compute. This treats HDL generation as state-space exploration rather than as a single-pass process. It allows the LLM to iteratively self-correct against strict logical constraints, effectively mirroring human debugging dynamics. Furthermore, to address the evaluation gaps identified in Section~\ref{sec:benchmark_ppa_gap}, validation must move beyond simple functional pass rates to system-level benchmarks capable of assessing optimization metrics. Finally, the effectiveness of an iterative evolutionary search depends on high-quality feedback. This feedback facilitates more accurate debugging, which in turn drives the evolutionary process toward faster convergence.

\section{Methodology}

Recognizing that the intricate nature of Verilog generation demands both extensive exploration and precise debugging, we architect two distinct algorithmic strategies: Idea-Guided Refinement (IGR) and Monte Carlo Tree Search (MCTS). The IGR strategy separates idea generation from implementation by incorporating internal knowledge from LLM or the referenced papers. MCTS utilizes tree-based search to navigate the design space. We select these independent strategies to target complementary optimization goals: IGR is employed to ensure global exploration of the architectural space to escape local optima for PPA optimization, while MCTS is chosen for precise exploitation to rigorously resolve local functional and timing constraints during Verilog code generation. This section formally defines the Verilog generation and optimization task in our framework, outlines the core algorithms and optimization objective, and presents the strategies that yield significant performance improvements.

\subsection{Problem Definition} \label{sec:problem_formulation}
We formulate the Verilog development process as a search optimization problem. The generation process explores a set of nodes $\mathcal{N} = \{N_1, N_2, \dots, N_n\}$. The state of the $i$-th node is defined as a tuple $N_i = (V_i, S_i, F_i)$, where $V_i$ is the candidate Verilog code, $S_i \in \mathbb{R}$ is the score, and $F_i$ is the diagnostic feedback.

The code $V_i$ is validated against a testbench $T$, which is defined as a set of distinct test cases $T = \{t_1, t_2, \dots, t_{|T|}\}$. An evaluator function $E$ executes these tests to derive the score and feedback:
\begin{equation}
    (S_i, F_i) = E(V_i, T)
    \label{eq:evaluate}
\end{equation}
Our framework utilizes an LLM to iteratively generate and refine these nodes to find an optimal design $N^*$:
\begin{equation}
    N^* = \underset{N_i \in \mathcal{N}}{\arg\max} \, S_i
\end{equation}
The definition of the scoring metric $S_i$ depends on the specific task objectives: \textbf{Generative Synthesis} or \textbf{PPA Optimization}.

For \textbf{Generative Synthesis}, the objective is strictly functional correctness. Let $p_i$ denote the number of test cases in $T$ successfully passed by $V_i$. The score is defined as the pass rate, with a large negative constant $C_{\text{penalty}}$ applied for compilation or simulation failures:
\begin{equation}
    S_{\text{gen}}(V_i) = 
    \begin{cases} 
    \frac{p_i}{|T|} & \text{if simulation success}, \\
    C_{\text{penalty}} & \text{otherwise}
    \end{cases}
    \label{eq:score_gen}
\end{equation}

For \textbf{PPA Optimization}, the objective is to minimize Power, Performance, and Area metrics while maintaining functional integrity. We formulate this as a constrained maximization problem where valid designs are scored based on the product of Area $A(V_i)$ and Latency $L(V_i)$, normalized by a scaling factor $\eta$:
\begin{equation}
    S_{\text{opt}}(V_i) = 
    \begin{cases} 
    -\frac{A(V_i) \cdot L(V_i)}{\eta} & \text{if } p_i = |T|, \\
    C_{\text{penalty}} & \text{otherwise}
    \end{cases}
    \label{eq:score_opt}
\end{equation}

Finally, the diagnostic feedback $F_i$ adapts dynamically to guide the LLM's next iteration:
\begin{equation}
    F_i = 
    \begin{cases} 
    \texttt{ErrorMsg} & \text{if } S_i = C_{\text{penalty}}, \\
    \texttt{Design Summary} & \text{else if Task is Gen}, \\
    \texttt{Opt Guidance} \oplus \texttt{Design Summary} & \text{else if Task is Opt}
    \end{cases}
\end{equation}

Equation~\ref{eq:score_opt} formulates the Area-Latency product as a rapid proxy for the evaluation function within our Verilog optimization framework. This approach offers two primary advantages. First, utilizing open-source tools such as Yosys and Iverilog yields a speedup of over $30\times$ compared to commercial workflows like Synopsys VCS and Design Compiler. Second, because area reduction is strongly correlated with power optimization, focusing on area allows for faster simulation while still achieving positive PPA outcomes. This eliminates the need to generate FSDB or VCD files for power evaluation, further accelerating the process. These improvements were validated using PrimeTime and the aforementioned commercial suite, as shown in Figure~\ref{fig:at_performance}.

\subsection{Frameworks}
\begin{figure}
    \centering
    \includegraphics[width=0.8\linewidth]{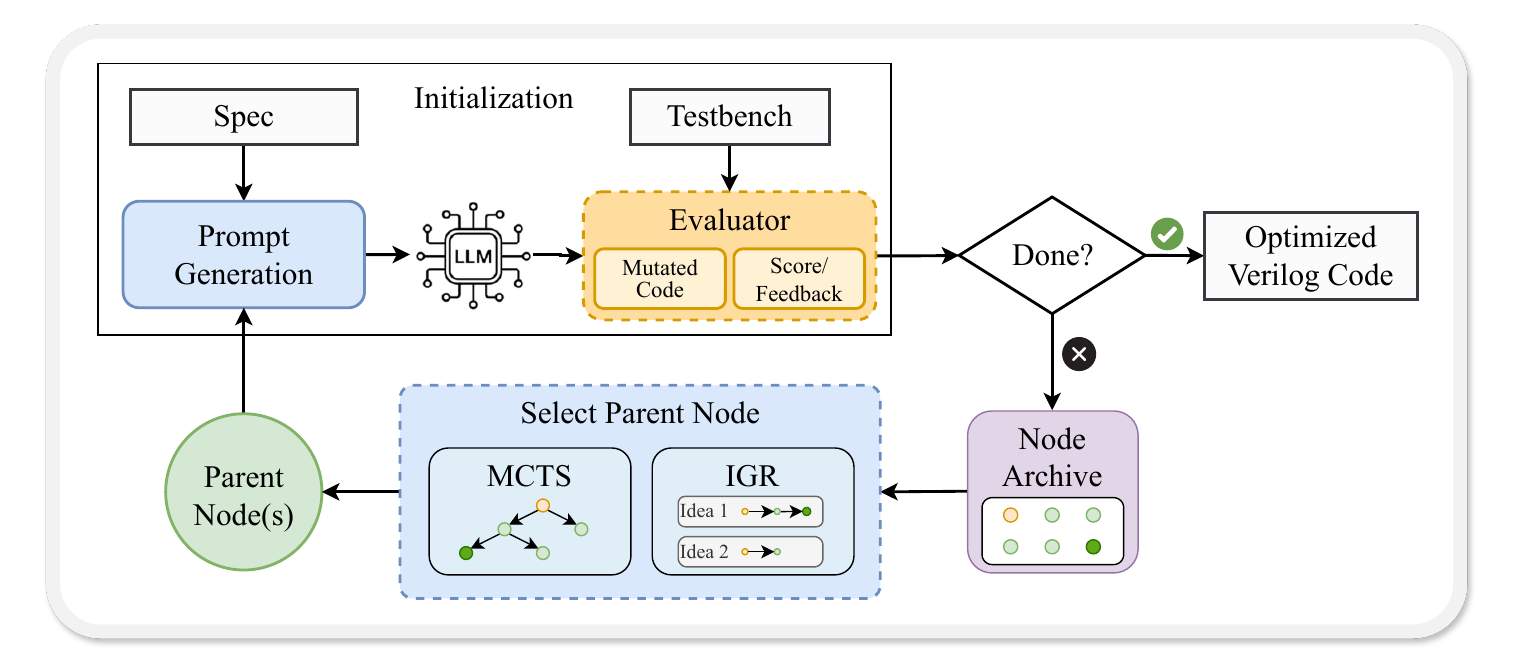}
    \caption{Overview of the EvolVE search framework.}
    \label{fig:general_algorithm}
\end{figure}

Our two evolutionary strategies share the unified framework illustrated in Figure~\ref{fig:general_algorithm} and detailed in Appendix~\ref{sec:general_framework}. Starting with a problem description, the framework prompts the LLM to generate the initial code. An evaluator assesses this code against a testbench to yield a quantitative score and textual feedback. As defined in Section~\ref{sec:problem_formulation}, a node comprises the code, score, and feedback. 
In the evolutionary phase, the parent node selection strategy retrieves a parent node from the archive by choosing one of two methods: IGR and MCTS. Using the parent node's code and feedback, the LLM produces a refined version. The cycle continues until a design passes all test cases ($S_\text{gen}=1$) in generative tasks, or until the maximum node limit is reached.

\begin{figure}
    \centering
    \includegraphics[width=1\linewidth]{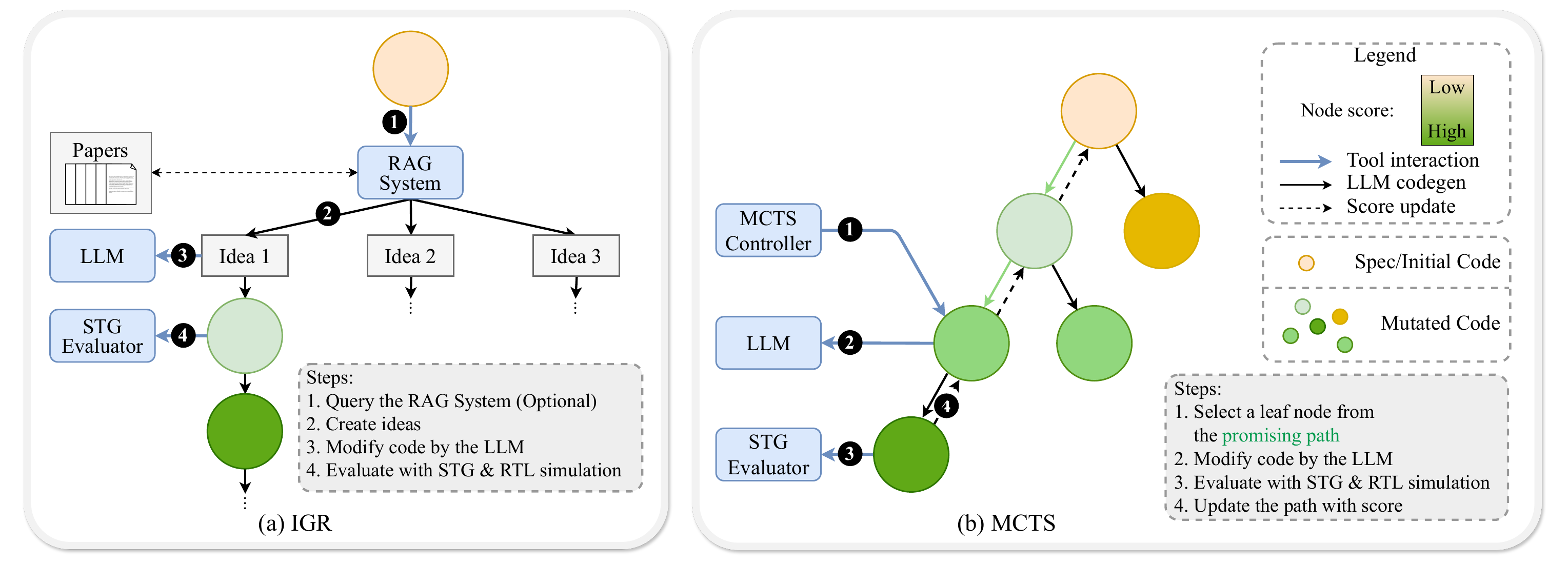}
    \caption{Comparative procedural flow for IGR and MCTS.}
    \label{fig:frameworks}
\end{figure}

\paragraph{Idea-Guided Refinement (IGR).}
Figure~\ref{fig:frameworks}(a) illustrates the IGR framework's multi-start strategy. We formalize this process into two distinct phases:

\begin{enumerate}
    \item \textbf{Idea Generation.} 
    The LLM first generates a set of high-level architectural concepts $\mathcal{I} = \{I_1, I_2, \dots, I_k\}$. This process leverages the LLM's intrinsic knowledge, optionally augmented by external insights drawn from referenced papers via Retrieval-Augmented Generation (RAG). Each idea $I_k$ is conditioned on the problem description and the history of preceding ideas $\{I_1, \dots, I_{k-1}\}$ to enforce diversity in the solution space.
    
    \item \textbf{Sequential Refinement.} 
    Each idea $I_k$ instantiates an independent refinement chain $C_k$. The chain is initialized with a root node $N_{k,1}$ derived from implementing $I_k$. We then apply a sequence of $m-1$ refinement steps. For each step $j \in \{2, \dots, m\}$, an Aider-style coder~\citep{aider} generates the child code $V_{k,j}$ by applying differential edits to the parent code $V_{k,j-1}$ based on the feedback $F_{k,j-1}$. This approach applies targeted, diff-based edits instead of regenerating the full file, thereby reducing LLM token consumption. Moreover, this single-chain approach aids sequential debugging through preserved, consistent code states across the context window. In contrast, applying such partial edits in MCTS is problematic, as managing multiple divergent children would saturate the context window. This yields a sequence of nodes $C_k = (N_{k,1}, \dots, N_{k,m})$. After exploring a total of $N_{\text{total}} = k \times m$ nodes across parallel chains, the framework selects the optimal node $N^*$ that maximizes the score $S$.
\end{enumerate}

\paragraph{Monte Carlo Tree Search (MCTS).}
Our framework employs Monte Carlo Tree Search (MCTS)~\citep{MCTS} to organize candidate nodes within a search tree $\mathcal{T}$, where the root node $N_{\text{root}}$ is initialized using LLM-generated Verilog code. Each node $N \in \mathcal{T}$ maintains a visit count $C(N)$ and a cumulative quality value $Q(N)$, which aggregates scores propagated from its child nodes via backpropagation. The search proceeds iteratively through the following steps:

\begin{enumerate}
    \item \textbf{Selection.}
    Starting from $N_{\text{root}}$, the algorithm recursively selects child nodes to traverse the tree until a leaf node $N_{\text{leaf}}$ is reached. At each step, the child node $N_{\text{child}}$ with the highest score based on a variant of the Upper Confidence Bound for Trees (UCT)~\citep{UCT} is selected. In practice, unexplored children are prioritized maximally to encourage exploration. The selection score \(UCT(N_{\text{parent}}, N_{\text{child}})\) is defined as follows:
    \begin{align}
    UCT(N_{\text{parent}}, N_{\text{child}}) &=
        \begin{cases}
            +\infty, & \text{if } C(N_{\text{child}}) = 0, \\
            \frac{Q(N_\text{child})}{C(N_{\text{child}})} + c \cdot \frac{\sqrt{\max(1, C(N_{\text{parent}}))}}{1 + C(N_{\text{child}})}, & \text{otherwise}
        \end{cases}
    \end{align}
    with \(c\) controlling the relative strength of the exploration term.

    \item \textbf{Code Generation and Evaluation.}
    We guide the code evolution using a design summary from the selected leaf node $N_{\text{leaf}}$. Incorporating this summary into the LLM prompt, alongside the parent's code $V_{\text{leaf}}$, enables improvement-focused generation of a new program $V_{\text{new}}$. Immediately following generation, $V_{\text{new}}$ is evaluated against the testbench to obtain a quality score $S_{\text{new}}$.
    
    \item \textbf{Expansion and Backpropagation.}
    Using the generated code and its evaluation score, a new node $N_{\text{new}}$ is created and inserted into the search tree $\mathcal{T}$ as a child of $N_{\text{leaf}}$. The node is assigned a depth value $\text{depth}(N_{\text{new}}) = \text{depth}(N_{\text{leaf}}) + 1$. Finally, the score $S_{\text{new}}$ is backpropagated up the tree to the root. For every ancestor node $N_a$ along the path, we update the statistics as follows:
    \begin{equation}
        C(N_a) \leftarrow C(N_a) + 1, \quad Q(N_a) \leftarrow Q(N_a) + S_{\text{new}}
    \end{equation}
\end{enumerate}

\subsection{Evaluation Mechanism: Structured Testbench Generation}

Existing benchmarks, such as VerilogEval, are limited by sparse input stimulus coverage and binary output feedback. This lack of evaluation fidelity obscures the distinction between near-correct solutions and fundamental failures. To address this, we introduce the Structured Testbench Generation (STG) mechanism, a deterministic verification engine designed to provide high-fidelity, dense rewards through three distinct phases.

\paragraph{Phase 1: Automated Signal Classification.} STG first parses the Design Under Test (DUT) port interface using a rigorous set of Regular Expressions to categorize signals into semantic groups: \textit{Clock/Reset}, \textit{Control}, and \textit{Datapath}. By targeting standard industry naming conventions (e.g., \texttt{clk}, \texttt{rst}, \texttt{valid}, \texttt{ready}), STG ensures deterministic, fast testbench instantiation without relying on LLM-based inference.

\paragraph{Phase 2: Scalable Hybrid Stimulus Generation.} To balance exhaustive coverage with simulation feasibility, STG employs a width-constrained stimulus strategy. For control signals with width $w \le 8$, STG generates exhaustive $2^w$ state-space coverage to verify all potential mode transitions. For wider control buses ($w > 8$), the system automatically switches to constrained-random sampling to prevent runtime explosion. Simultaneously, for datapath buses, STG utilizes optimized random sampling seeded with corner cases (e.g., zero, max-value, alternating bits). The resulting testbenches adhere to Standard IEEE 2005 Verilog, making the STG simulator-agnostic and fully compatible with open-source tools like Icarus Verilog as well as commercial simulators like VCS.

\paragraph{Phase 3: Fine-Grained Functional Gradient.} STG enforces strict temporal alignment between the DUT and the golden reference model. Checks are triggered at critical state transitions, such as post-reset and after a brief delay following clock edges, to isolate sequential errors. Crucially, unlike prior works that rely on sparse bandit feedback, STG calculates a continuous fine-grained correctness score $P_{stg} \in [0, 1]$, which is defined as the pass rate across the generated test vectors. This acts as a functional gradient, allowing the search engine to prioritize candidates that are partially correct, thereby accelerating algorithmic convergence in complex design spaces.
\section{Experiments}

\subsection{Experimental Setup}
\label{sec:exp_setup}

We evaluate our framework using three self-hosted models, Siliconmind-7B (detailed in Appendix~\ref{sec:siliconmind-7b}), GPT-OSS-120B~\citep{gptoss}, and DeepSeek-R1-FP4~\citep{deepseekr1}, alongside the proprietary Gemini-2.0-Flash. These models are assessed across three generative benchmarks focusing on functional correctness: VerilogEval v2 \citep{verilogeval2}, Mod-VerilogEval v2, and RTLLM v2 \citep{rtllm}. Additionally, we evaluate PPA optimization capabilities using our proposed IC-RTL benchmark.

Our evaluation toolchain integrates both open-source and commercial ecosystems. We employ Icarus Verilog (Iverilog) \citep{iverilog} and Yosys \citep{yosys} for rapid simulation and synthesis, alongside the Synopsys suite (VCS, Design Compiler, and PrimeTime) for rigorous validation. Underlying this framework, our self-hosted models operate on a cluster of 8 NVIDIA H100 GPUs. Specifically, DeepSeek-R1-FP4 leverages tensor and expert parallelism within its Mixture-of-Experts (MoE) architecture, while Siliconmind-7B utilizes pure data parallelism.

Regarding hyperparameters, we standardize the evaluation budget to 300 nodes across all experiments. We configure the LLM temperature to $0.6$ and the penalty constant $C_{\text{penalty}}$ to $-10^5$. For specific search strategies, IGR is set to generate $k=60$ initial ideas refined over $m=5$ steps, while MCTS employs an expansion rate of 3 child nodes and an exploration constant of $c=1.4$.

\subsection{Verilog Code Generation}
\begin{table}[t]
    \centering
    \begin{threeparttable}
    \caption{Performance comparison on the original VerilogEval v2 and RTLLM v2. The Pass columns show the Pass rate (\%) progression at computational node budgets of \textbf{15 / 50 / 100 / 200 / 300}.}
    \label{tab:main-results}
    
    \begin{tabular}{@{}lcc@{}}
        \toprule
        \multirow{2}{*}{\textbf{Method}} & \multicolumn{1}{c}{\textbf{VerilogEval v2}} & \multicolumn{1}{c}{\textbf{RTLLM v2}} \\
        \cmidrule(lr){2-2} \cmidrule(lr){3-3}
         & \textbf{Pass (\%)} & \textbf{Pass (\%)} \\
        \midrule

        \rowcolor{lightgray!50} \multicolumn{3}{c}{\textit{Baselines (Pass@15 / 50 / 100 / 200 / 300)}} \\
        \midrule
        Gemini-2.0-Flash\textsuperscript{\dag} & 
        72.4 / 74.4 / 77.6 / 78.8 / 80.1 & 
        62.0 / 72.0 / 76.0 / 78.0 / 78.0 \\

        Siliconmind-7B & 
        82.1 / 87.8 / 89.7 / 91.7 / 91.7 & 
        76.0 / 82.0 / 84.0 / 84.0 / 84.0 \\

        GPT-OSS-120B & 
        92.9 / 93.6 / 94.2 / 94.2 / 94.2 & 
        76.0 / 76.0 / 76.0 / 76.0 / 76.0 \\
        
        DeepSeek-R1-FP4 & 
        91.0 / 94.2 / 94.9 / 94.9 / 95.5 & 
        82.0 / 84.0 / 84.0 / 86.0 / 86.0 \\
        
        \midrule
        \rowcolor{lightgray!50} \multicolumn{3}{c}{\textit{Previous Work}} \\
        \midrule
        REvolution (Deepseek-V3 w/ 200 nodes)\textsuperscript{\dag} & 95.5 & 88.0 \\
        REvolution (Llama-3.3-70B w/ 200 nodes)\textsuperscript{\dag} & 88.5 & 84.0 \\
        VerilogCoder (Llama-3-70B w/ 100 nodes) & 67.3 & - \\
        VerilogCoder (GPT-4-Turbo w/ 100 nodes)\textsuperscript{\dag} & 94.2 & - \\

        \midrule
        \rowcolor{lightgray!50} \multicolumn{3}{c}{\textit{Our Work w/o STG (Nodes: 15 / 50 / 100 / 200 / 300)}} \\
        \midrule

        Gemini-2.0-Flash-IGR\textsuperscript{\dag} & 
        76.3 / 85.3 / 86.5 / 86.5 / 86.5 & 
        66.0 / 70.0 / 74.0 / 80.0 / 80.0 \\
        
        Gemini-2.0-Flash-MCTS\textsuperscript{\dag}& 
        77.6 / 81.4 / 84.6 / 86.5 / 87.8 & 
        68.0 / 78.0 / 80.0 / 82.0 / 82.0 \\
        
        Siliconmind-7B-MCTS & 
        84.0 / 85.3 / 86.5 / 90.4 / 92.3 & 
        76.0 / 80.0 / 82.0 / 84.0 / 88.0 \\

        GPT-OSS-120B-IGR & 
        89.1 / 95.5 / 96.2 / 96.8 / 96.8  & 
        80.0 / 84.0 / 88.0 / 90.0 / 90.0 \\

        GPT-OSS-120B-MCTS & 
        89.7 / 93.6 / 94.9 / 95.5 / 96.2 & 
        78.0 / 84.0 / 86.0 / 88.0 / 88.0 \\

        \textbf{DeepSeek-R1-FP4-MCTS} & 
        \textbf{94.2} / \textbf{96.8} / \textbf{96.8} / \textbf{97.4} / \textbf{98.1} & 
        82.0 / 84.0 / \textbf{86.0} / \textbf{90.0} / \textbf{92.0} \\

        \bottomrule
    \end{tabular}
    
    \begin{tablenotes}[flushleft]
        \footnotesize 
        \item \textsuperscript{\dag} Denotes models accessed via API calling.
    \end{tablenotes}
    \end{threeparttable}
\end{table}

\paragraph{Performance Comparison.}
In this section, we quantitatively analyze the efficacy of the EvolVE framework across varying LLM architectures. By benchmarking against established baselines, we show that our inference-time search strategy consistently augments the base capabilities of the underlying models.

Table~\ref{tab:main-results} presents a comprehensive quantitative evaluation of the original VerilogEval v2 and RTLLM v2 benchmarks. Our results demonstrate that inference-time search significantly augments the design generation capabilities of LLMs, consistently outperforming both standard baselines and prior state-of-the-art methodologies. The most notable performance is observed with the DeepSeek-R1-FP4-MCTS configuration. Even under a highly constrained budget of 15 computational nodes, this model achieves a pass rate of 94.2\% on VerilogEval v2, effectively matching the best reported performance of VerilogCoder (GPT-4-Turbo w/ 100 nodes). As the computational budget scales to 300 nodes, DeepSeek-R1-FP4-MCTS achieves peak pass rates of 98.1\% on VerilogEval v2 and 92.0\% on RTLLM v2. This represents a substantial improvement over the REvolution framework, which achieved 95.5\% using Deepseek-V3, validating the superiority of our MCTS implementation in navigating complex design spaces.

We further observe performance gains in the commercial model API. For Gemini-2.0-Flash, the standard baseline notably saturates at 80.1\% on VerilogEval v2. However, by applying our evolutionary strategies, we successfully unlock latent reasoning capabilities, boosting performance to 86.5\% with IGR and 87.8\% with MCTS. While both approaches drive significant gains, MCTS yields higher stability and superior peak performance compared to IGR across most architectures.

\begin{figure}[t]
    \centering
    \includegraphics[width=1.0\linewidth]{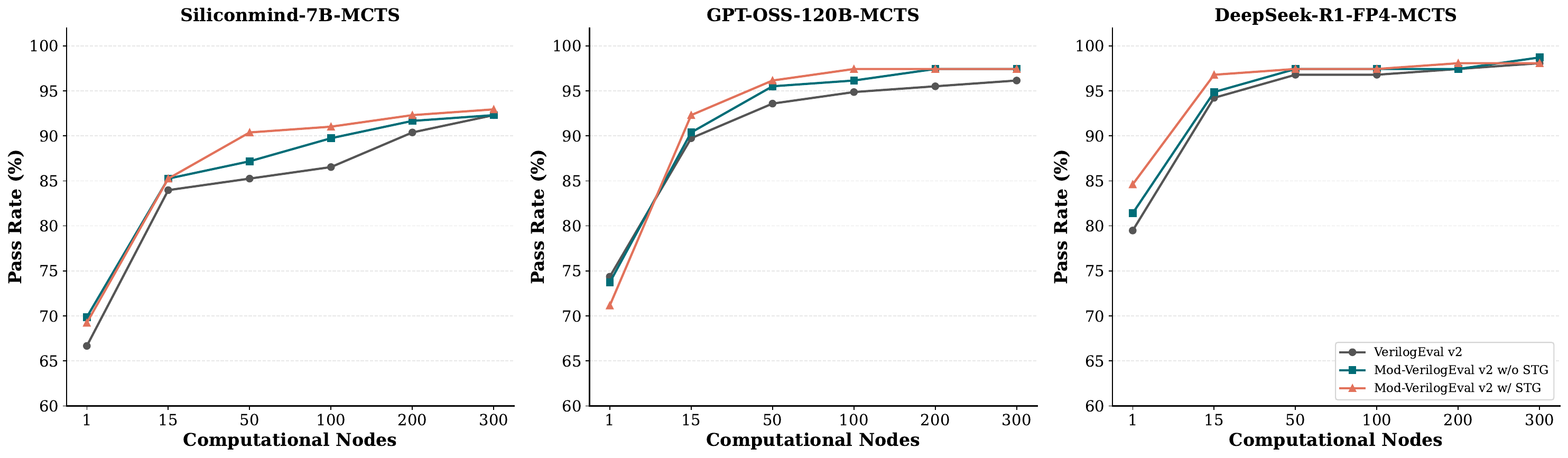}
    \caption{Performance scaling of MCTS across benchmarks and models.}
    \label{fig:results-scaling}
\end{figure}

\paragraph{Scalability and Computational Budget.} Figure~\ref{fig:results-scaling} illustrates the performance scaling trajectories of our MCTS approach across varying computational budgets (from 1 to 300 nodes). Unlike standard baselines, which often plateau after a few attempts, our framework continues to extract performance gains as the node count increases. We observe distinct scaling behaviors contingent on the base model's reasoning capacity. For highly capable models like DeepSeek-R1-FP4, the search is exceptionally sample-efficient. As detailed in our collected data, this configuration resolves 84.6\% of the benchmark problems at the very first node and requires an average of only 2.49 nodes to reach a solution. Conversely, for models with lower baseline pass rates, the MCTS framework effectively utilizes the expanded budget address the subset of high-difficulty problems. This is best exemplified by the Siliconmind-7B-MCTS experiments. While the model solves the majority of easier tasks early, it continues to harvest solutions deep into the search tree. Specifically, Siliconmind-7B-MCTS demonstrates a steady climb, rising from 85.3\% (15 nodes) to 92.9\% (300 nodes). This monotonic increase confirms that the framework does not merely perform rejection sampling, but actively utilizes additional compute to correct subtle logic errors in hard-to-solve benchmarks.

\begin{figure}[t]
    \centering
    \includegraphics[width=0.9\linewidth]{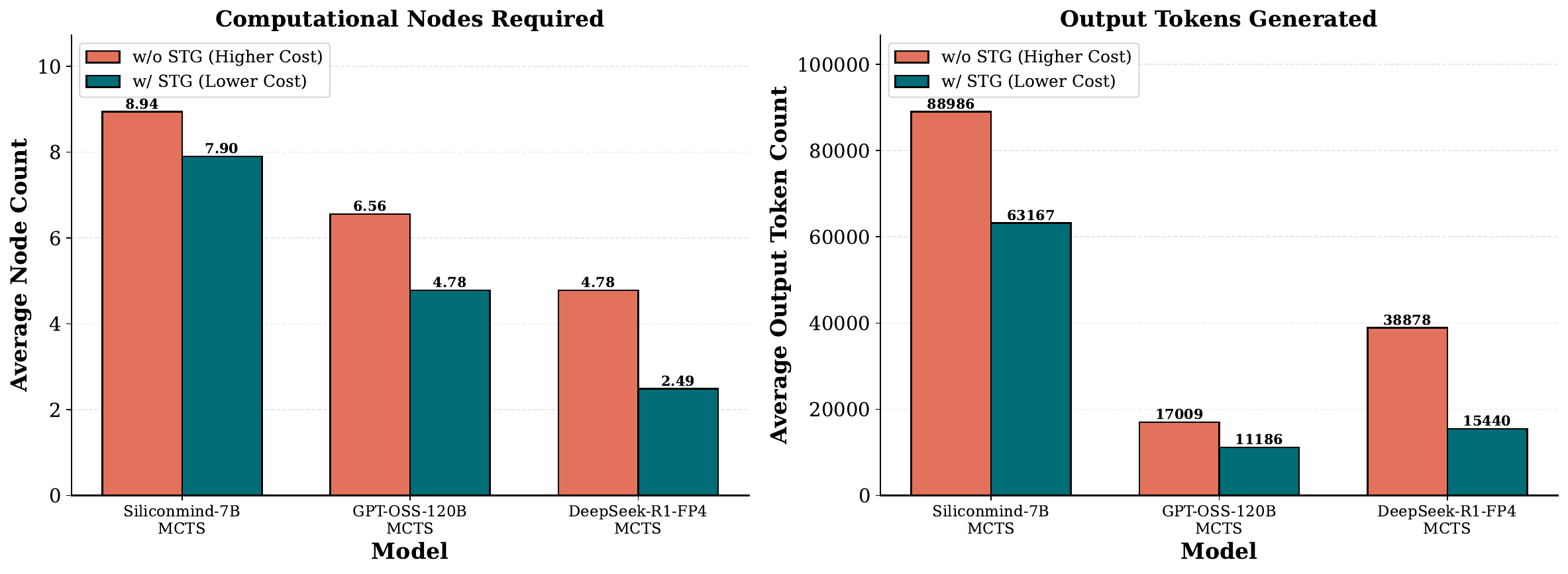}
    \caption{Convergence efficiency gains from STG on Mod-VerilogEval v2 benchmark.}
    \label{fig:stg_efficiency}
\end{figure}

\paragraph{Efficacy of Structured Testbench Generation.} To quantify the specific contribution of the STG module, we conducted ablation studies comparing the framework's behavior with and without syntax guidance. As illustrated in Figure~\ref{fig:results-scaling}, the STG-enhanced configuration maintains performance parity with the unguided variant at the upper bound of the computational budget (300 nodes). However, the decisive advantage of STG is revealed in Figure~\ref{fig:stg_efficiency}, which highlights a dramatic improvement in convergence efficiency. By strictly enforcing syntactic validity during the search, STG significantly lowers the computational overhead required to reach these solutions. For the DeepSeek-R1-FP4 architecture, the average node count required to resolve the Mod-VerilogEval v2 benchmark drops from 4.78 to 2.49. Furthermore, the total output token consumption decreases by over 60\% (from 38,818 to 15,440). These metrics confirm that while STG achieves comparable asymptotic accuracy, it optimizes the search trajectory, allowing the framework to attain target performance levels with substantially reduced latency and inference cost.
\subsection{PPA Optimization}

\paragraph{Benchmark: IC-RTL.}
To rigorously evaluate the framework's capability in handling complex, industry-relevant hardware logic, we introduce IC-RTL, a suite comprising hand-crafted design problems and advanced tasks derived from the Taiwan National IC Design Contest \citep{TaiwanICContest}. Unlike existing benchmarks that prioritize simple syntax at limited scales, IC-RTL demands the implementation of specific algorithmic structures and memory optimization techniques to achieve high-performance targets. Crucially, these designs are compatible with both commercial and open-source platforms and offer a significant design space for PPA optimization. The constituent tasks are detailed below:

\begin{itemize}
    \item \textbf{Local Binary Patterns (LBP):}
    This module computes texture descriptors for a $128\times128$ grayscale image using a sliding $3\times3$ window. The core challenge lies in minimizing memory access latency. An optimized design typically employs line buffering to store previous rows locally, enabling the concurrent processing of pixel neighborhoods without redundant fetches from main memory.
    
    \item \textbf{Systolic Array (General Matrix Multiplication):} 
    This task involves designing a highly parallel matrix multiplication engine using a systolic architecture. The key optimization objective is data reuse and synchronization. The design must orchestrate a precise data flow where operand tiles are broadcast across an array of Processing Elements (PEs), maximizing throughput while minimizing I/O bandwidth. Specifically, the reported latency captures only the dataflow and execution time of the systolic engine.
    
    \item \textbf{Image Convolutional Circuit (CONV):} 
    This task implements a hardware accelerator for a Convolutional Neural Network (CNN) layer, including zero-padding, $3\times3$ convolution, bias addition, ReLU activation, and $2\times2$ max-pooling. A significant constraint is the handling of fixed-point arithmetic (Q4.16) and the management of intermediate data dependencies between the convolution and pooling layers.
    
    \item \textbf{Job Assignment Machine (JAM):} 
    This module solves a combinatorial optimization problem to minimize the cost of assigning 8 workers to 8 jobs. Since the problem requires exploring all $8!$ ($40,320$) permutations, the hardware must implement an efficient lexicographic permutation generation algorithm alongside a pipelined cost-accumulation path to meet timing constraints.
    
    \item \textbf{Huffman Coding (HC):} 
    This task requires the hardware generation of variable-length Huffman codes based on symbol frequency statistics. The complexity lies in implementing sorting and tree-construction algorithms in hardware. An optimized implementation typically utilizes parallel sorting networks or priority queue structures to perform the combine-and-split phases within limited clock cycles.
    
    \item \textbf{Distance Transform (DT):} 
    The engine computes the chessboard distance for a binary image using a two-pass algorithm (Forward and Backward passes). The optimization challenge involves managing read-after-write dependencies where the backward pass depends on the results of the forward pass, requiring efficient memory arbitration and pipeline stalling mechanisms to prevent data hazards.

\end{itemize}

\paragraph{Optimization Metrics.}
For the PPA optimization task (Equation~\ref{eq:score_opt}), the normalization factor is set to $\eta = 10^5$. This scaling ensures that the scores for functionally valid designs remain strictly greater than the penalty threshold, facilitating effective gradient discrimination during search.

\paragraph{PPA Performance Analysis.}

\begin{figure}
    \centering
    \includegraphics[width=0.9\linewidth]{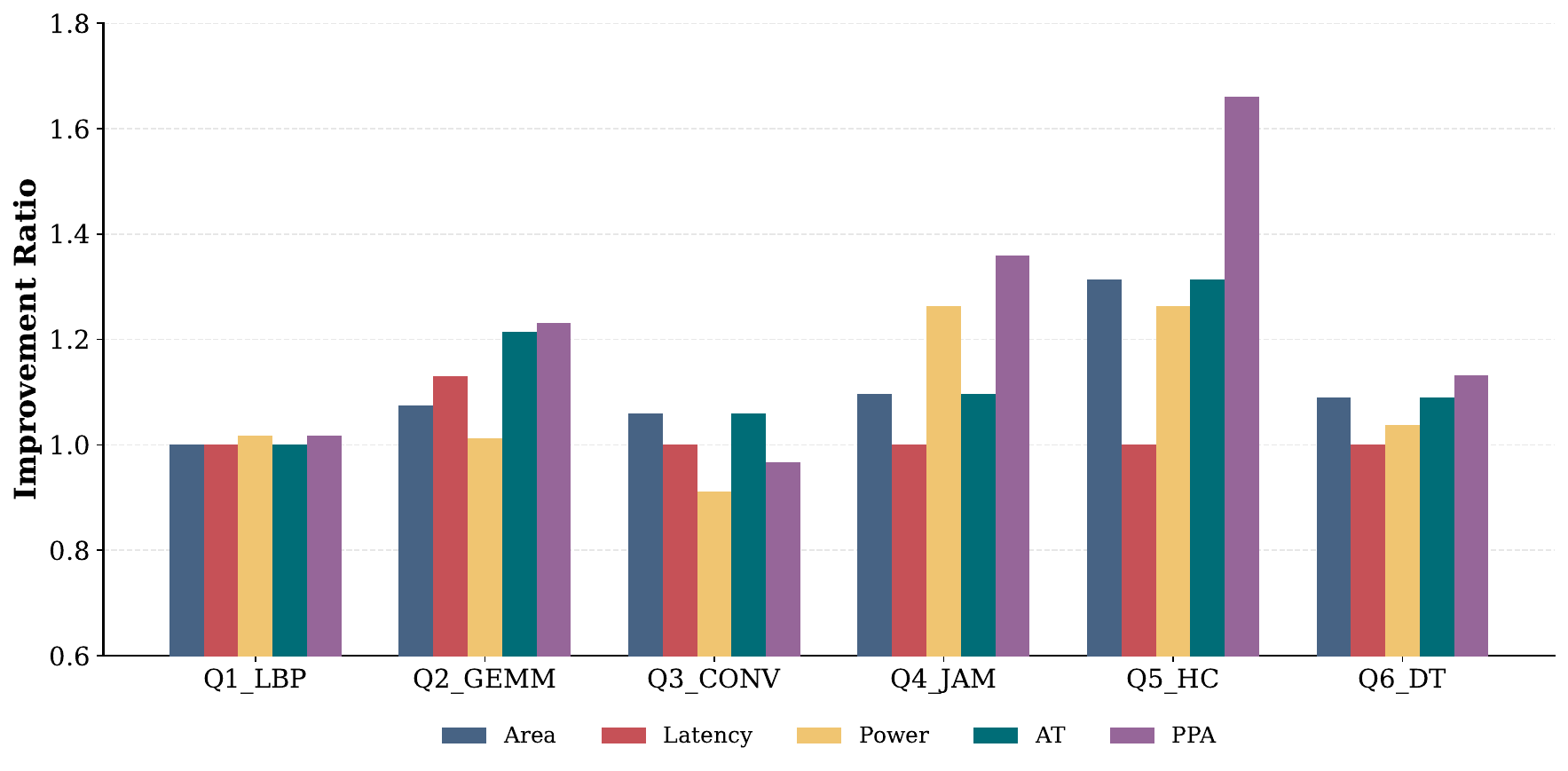}
    \caption{PPA improvements relative to human baselines on the IC-RTL benchmark.}
    \label{fig:ppa_performance}
\end{figure}

Figure~\ref{fig:ppa_performance} highlights the capability of the EvolVE framework on the IC-RTL benchmark through the Idea-Guided Refinement (IGR) strategy utilizing the DeepSeek-R1-FP4 model. To validate the results in a realistic setting, all generated RTL designs are synthesized using Synopsys Design Compiler under the TSMC 180nm technology node. The chart reports the improvement ratio normalized against baseline solutions, where a value greater than 1.0 indicates superior efficiency.

Crucially, this evaluation sweeps discrete target clock periods ($T_{clk} \in \{3, \dots, 7\}$ $ns$) and defines latency as the product of simulation cycles and the clock period. We prioritize this methodology over measuring timing slack for two reasons. First, commercial synthesis tools are constraint-driven: they fundamentally alter the circuit topology (e.g., selecting different standard cell drive strengths or arithmetic architectures) based on the specific clock target. Consequently, simply measuring slack on a design synthesized with a relaxed clock is meaningless, as it fails to capture the structural optimizations the tool would perform under a tighter constraint. Second, the global optimum for PPA efficiency rarely aligns with the maximum achievable frequency. By optimizing under discrete clock constraints, we identify the specific operating point that maximizes the Area-Latency (AT) product, rather than forcing the tool to close timing at theoretical limits, which incurs disproportionate area and power penalties. 

Figure~\ref{fig:ppa_performance} compares the optimized and initial implementations under identical operation frequency, with full details provided in Appendix~\ref{sec:ic_rtl_detailed}. The results demonstrate consistent optimization gains across diverse architectural patterns:

\begin{itemize}

\item \textbf{Baseline Convergence (Q1\_LBP):} The framework converged to a design that matches the manual baseline's performance profile. The optimized latency and area metrics remain within 1\% of the initial implementation, indicating that the human-designed Local Binary Pattern kernel effectively occupies a local optimum within the explored design space.

\item \textbf{Systolic Array Efficiency (Q2\_GEMM):} The Matrix Multiplication task demonstrates the framework's strength in optimizing regular dataflows. The design achieved a 12\% reduction in Latency while maintaining neutral Area and lower Power. This gain stems from the optimizer effectively rescheduling the systolic array's control logic to maximize data reuse and reduce pipeline bubbles, directly improving the Area-Latency (AT) product without incurring hardware overhead.

\item \textbf{Area-Power Trade-off (Q3\_CONV):} For the Convolution kernel, the framework executed a targeted trade-off: it reduced Area by 6\% to improve spatial efficiency while maintaining fixed Latency. However, this compaction required higher switching activity, leading to a Power regression (0.45 mW to 0.50 mW) that neutralized the overall PPA gain. 

\item \textbf{Area-Power Co-Optimization (Q4\_JAM):} The Job Assignment Machine task highlights a strategy prioritizing Area efficiency. The framework achieved significant reductions in both Area (31\%) and Power (26\%) without sacrificing latency, driving a 36\% composite PPA improvement. This confirms the model's ability to identify complex optimization paths that aggressively minimize the number of logic gates while simultaneously maximizing energy efficiency. It further demonstrates that the draft of the AT product is effectively in the final PPA optimization.

\item \textbf{Sorting Network Optimization (Q5\_HC):} The impressive gain refers to the complex design space for the Huffman Coding task, where the framework optimized the complex decision trees and sorting networks. It delivered a 66\% peak PPA improvement, driven by simultaneous reductions in Area (30\%) and Power (25\%). This confirms the model's ability to efficiently restructure irregular control-flow logic.

\item \textbf{Datapath Scaling (Q6\_DT):} The Distance Transform task exhibits improvements across Area and Power metrics, resulting in a robust 13\% PPA improvement, indicating our framework's ability to handle different kinds of complex problems.

\end{itemize}

\begin{figure}
    \centering
    \includegraphics[width=0.9\linewidth]{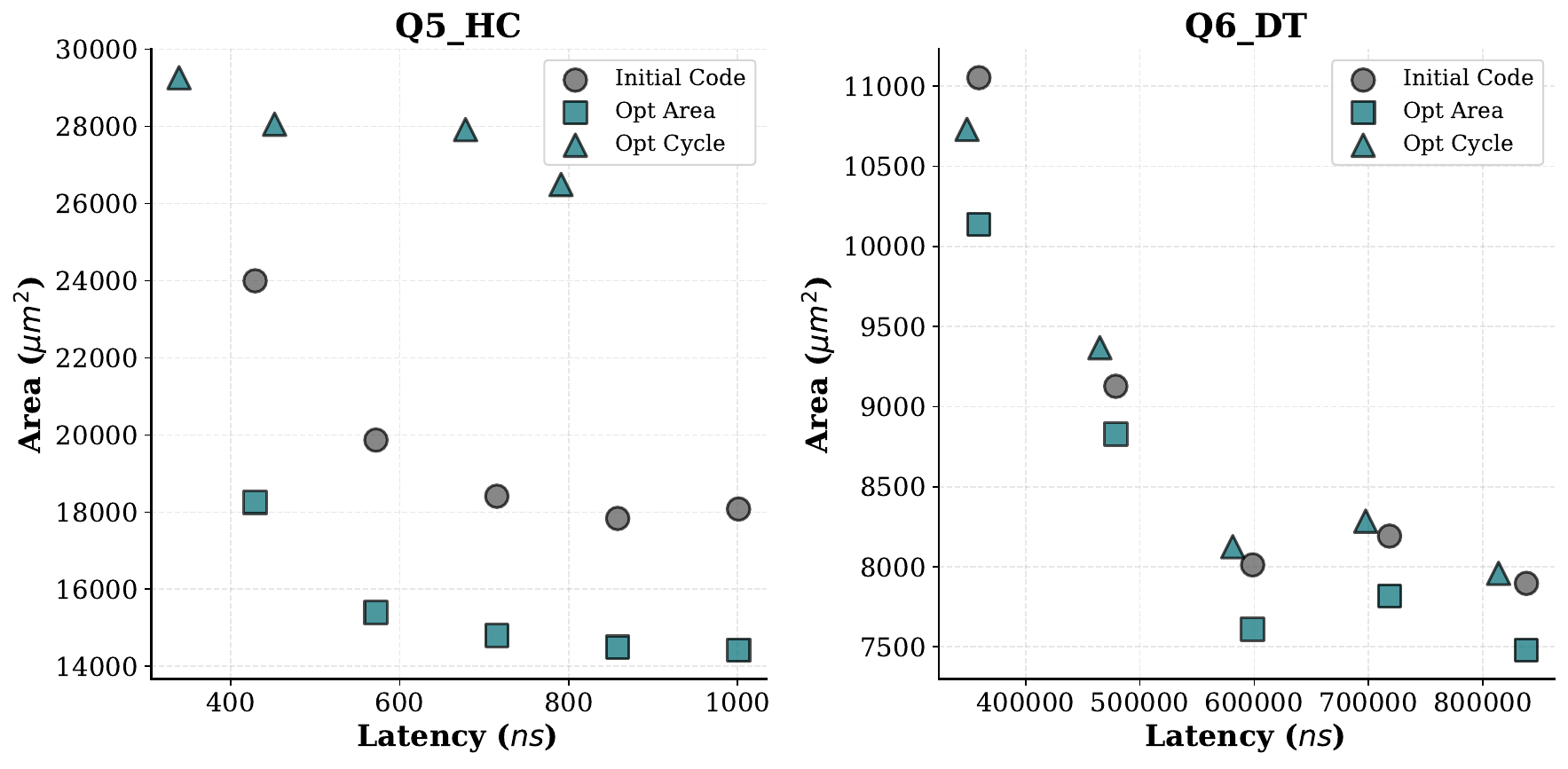}
    \caption{Area and Latency evaluation on the IC-RTL benchmark.}
    \label{fig:at_performance}
\end{figure}

\paragraph{Oriented Design Space Exploration}

Beyond generalized scalar improvements, 
Figure~\ref{fig:at_performance} demonstrates the framework's capability to steer optimization trajectories toward specific constraints. Using the Huffman Coding (Q5\_HC) and Distance Transform (Q6\_DT) tasks, we directed the agent to generate distinct architectural variants by injecting targeted prompts: one prioritizing spatial efficiency (\texttt{Opt Area}) and another prioritizing temporal speed (\texttt{Opt Cycle}). Sweeping across clock periods ($T_{clk} \in \{3, \dots, 7\}$ $ns$), the results in Figure~\ref{fig:at_performance} reveal divergent optimization behaviors that effectively populate the Pareto frontier:

\begin{itemize}\item \textbf{Area-Centric Optimization (\texttt{Opt Area}):} As shown by the square markers in Figure~\ref{fig:at_performance} (left), this strategy strictly enforces the baseline latency constraints while aggressively minimizing logic utilization. For Q5\_HC, this resulted in a 31\% reduction in Area (from $2.39 \times 10^4$ to $1.82 \times 10^4$ $\mu m^2$ at $3ns$) without degrading performance. The vertical alignment between the baseline (circles) and area-optimized (squares) points confirms that the framework successfully isolated area savings without impacting timing closure.

\item \textbf{Latency-Centric Optimization (\texttt{Opt Cycle}):} Represented by the triangle markers, this variant successfully decoupled latency from the standard clock scaling. For Q5\_HC, the framework achieved a 26\% reduction in execution time (lowering latency from $429ns$ to $339ns$ at $3ns$ clock). Crucially, this speedup incurred a necessary area penalty (increasing from 2.4 $\times 10^4$ to 2.9 $\times 10^4$ $\mu m^2$), demonstrating the framework's ability to navigate the classic Area-Delay trade-off to deliver high-performance solutions when requested.
\end{itemize}

This divergence confirms that EvolVE does not merely converge to a single local optimum but allows designers to negotiate the PPA trade-off space dynamically. Whether the goal is minimizing the number of logic gates for IoT applications or maximizing throughput for high-performance computing, the framework enables specialized, constraint-aware implementation.
\subsection{Ablation Study}
\paragraph{Case Study: Architectural Evolution in GEMM.}
To illustrate the depth of optimization, we analyze the results for Q2\_GEMM, where the framework simultaneously improved both area and cycle counts. The baseline human design implements an output-stationary systolic array that handles variable matrix dimensions by decomposing them into $4 \times 4$ matrix multiplication operations. This architecture includes dedicated input and weight reshape buffers to align data for the systolic flow, necessitating a flattened port interface due to limitations in Icarus Verilog.

\begin{itemize}
    \item \textbf{Micro-Architectural Refinement:} We found that the framework successfully optimized the data buffering scheme. For $n \times n$ matrix multiplication, the baseline design utilized a  $(2n - 1)$-row buffer to handle the parallelogram data shaping required for the systolic array. The optimized design compressed this to n rows by employing intelligent multiplexing to select data. Additionally, the framework eliminated redundant buffer registers among PE arrays and the input buffer and retimed the critical path. Under a $4ns$ clock constraint, this reduced the total latency from $1448ns$ to $1280ns$, while simultaneously reducing area from 339,266 to 315,770 $\mu m^2$.
    \item \textbf{Novel Architectural Discovery:} Through extended evolutionary search, the framework autonomously evolved the design from a pure output-stationary model to a weight-output stationary hybrid, reducing latency from $1280ns$ to $776ns$. This transformation eliminated separate weight buffers in favor of direct weight insertion and broadcast inputs for the operand matrix. This result is significant as it validates the framework's ability to derive architectural-level optimization, transcending local RTL fixes to discover superior high-level topologies.
    
\end{itemize}

\begin{table}
    \centering
    \caption{Detailed performance comparison for GEMM candidate programs.}
    \label{tab:gemm_comparison}
    \begin{tabular}{@{} lccccc @{}}
        \toprule
        \textbf{Candidate} & \textbf{Area} & \textbf{Clock} & \textbf{Cycle} & \textbf{Timing} & \textbf{AT} \\
        (Best Area*Time) & ($\mu m^2$) & ($ns$) & & ($ns$) & (Area * Timing) \\
        \midrule
        Initial Program  & 269657 & 4 & 362 & 1448 & 390463336 \\
        Dolphin's Program & 257906 & 4 & 320 & 1280 & 330119680 \\
        \bottomrule
    \end{tabular}
\end{table}

\section{Conclusion} In this work, we present EvolVE, a unified, model-agnostic evolutionary framework that enables LLMs to generate and optimize high-quality, functionally correct Verilog code directly from natural language specifications. By instantiating two distinct search strategies, Idea-Guided Refinement (IGR) and Monte Carlo Tree Search (MCTS), we provide the first study establishing MCTS as superior for functional generation and IGR as optimal for PPA optimization. Another innovation, the Structured Testbench Generation (STG) engine, replaces coarse-grained feedback with fine-grained scores, significantly accelerating convergence and enabling the framework to solve complex logic problems where baseline models saturate. Furthermore, we introduce Mod-VerilogEval v2 to address inconsistencies in existing standards, and IC-RTL, a benchmark suite that bridges the gap between academic datasets and industry-level complexity, providing a rigorous platform for evaluating design optimization capabilities.

Extensive experiments across VerilogEval v2, Mod-VerilogEval v2, RTLLM v2, and IC-RTL demonstrate state-of-the-art performance, with the framework reaching a 98.1\% pass rate using limited evaluation budgets. The approach proves effective across all model scales, notably boosting the accuracy of the Siliconmind-7B model from 82.1\% to over 92\%. In terms of optimization, our framework outperforms human-designed references, reducing the PPA product by up to 66\% on Q5 and 17\% in the geometric mean across the IC-RTL suite. Crucially, we observe that EvolVE can actively steer optimization trajectories and autonomously discover architectural transformations. These results suggest that when augmented with evolutionary guidance, LLMs can transcend the limitations of model scale and data scarcity, evolving into active agents capable of specification to RTL generation, targeted optimization, and microarchitectural design space exploration.

Despite these advances, several limitations remain. Currently, STG relies on the availability of executable reference models (golden designs in C or Verilog) to verify signal correctness, which may not always be available for novel IP. Additionally, benchmarks specifically targeted at PPA optimization remain scarce, and integrating fully automated synthesis flows to provide real-time PPA feedback requires further development.

Future work will focus on two key directions. First, inspired by methodologies like ACE~\citep{ace}, we aim to integrate a formal knowledge base (playbook) of microarchitectural optimization patterns to guide PPA-driven exploration. Second, we plan to leverage the exploratory nature of MCTS to better optimize for power, performance, and area, rather than functional correctness alone. We believe these directions will help transform RTL design from a manual, expert-limited process into an automated, systematic discipline, driving a fundamental shift in the IC industry.
\bibliographystyle{apalike}
\bibliography{references}

@article{verilogeval2,
  title={Revisiting verilogeval: Newer llms, in-context learning, and specification-to-rtl tasks},
  author={Pinckney, Nathaniel and Batten, Christopher and Liu, Mingjie and Ren, Haoxing and Khailany, Brucek},
  journal={arXiv e-prints},
  pages={arXiv--2408},
  year={2024}
}

@article{iverilog,
  title={Icarus verilog: open-source verilog more than a year later},
  author={Williams, Stephen and Baxter, Michael},
  journal={Linux Journal},
  volume={2002},
  number={99},
  pages={3},
  year={2002},
  publisher={Belltown Media Houston, TX}
}

@article{alphaevolve,
  title={AlphaEvolve: A coding agent for scientific and algorithmic discovery},
  author={Novikov, Alexander and V{\~u}, Ng{\^a}n and Eisenberger, Marvin and Dupont, Emilien and Huang, Po-Sen and Wagner, Adam Zsolt and Shirobokov, Sergey and Kozlovskii, Borislav and Ruiz, Francisco JR and Mehrabian, Abbas and others},
  journal={arXiv preprint arXiv:2506.13131},
  year={2025}
}

@article{rtlcoder,
  title={Rtlcoder: Fully open-source and efficient llm-assisted rtl code generation technique},
  author={Liu, Shang and Fang, Wenji and Lu, Yao and Wang, Jing and Zhang, Qijun and Zhang, Hongce and Xie, Zhiyao},
  journal={IEEE Transactions on Computer-Aided Design of Integrated Circuits and Systems},
  year={2024},
  publisher={IEEE}
}

@article{scalertl,
  title={ScaleRTL: Scaling LLMs with Reasoning Data and Test-Time Compute for Accurate RTL Code Generation},
  author={Deng, Chenhui and Tsai, Yun-Da and Liu, Guan-Ting and Yu, Zhongzhi and Ren, Haoxing},
  journal={arXiv preprint arXiv:2506.05566},
  year={2025}
}

@article{codev,
  title={CodeV-R1: Reasoning-Enhanced Verilog Generation},
  author={Zhu, Yaoyu and Huang, Di and Lyu, Hanqi and Zhang, Xiaoyun and Li, Chongxiao and Shi, Wenxuan and Wu, Yutong and Mu, Jianan and Wang, Jinghua and Zhao, Yang and others},
  journal={arXiv preprint arXiv:2505.24183},
  year={2025}
}

@article{symrtlo,
  title={SymRTLO: Enhancing RTL Code Optimization with LLMs and Neuron-Inspired Symbolic Reasoning},
  author={Wang, Yiting and Ye, Wanghao and Guo, Ping and He, Yexiao and Wang, Ziyao and Tian, Bowei and He, Shwai and Sun, Guoheng and Shen, Zheyu and Chen, Sihan and others},
  journal={arXiv preprint arXiv:2504.10369},
  year={2025}
}

@article{veriopt,
  title={Veriopt: Ppa-aware high-quality verilog generation via multi-role llms},
  author={Tasnia, Kimia and Garcia, Alexander and Farheen, Tasnuva and Rahman, Sazadur},
  journal={arXiv preprint arXiv:2507.14776},
  year={2025}
}

@article{rtlsquad,
  title={Rtlsquad: Multi-agent based interpretable rtl design},
  author={Wang, Bowei and Xiong, Qi and Xiang, Zeqing and Wang, Lei and Chen, Renzhi},
  journal={arXiv preprint arXiv:2501.05470},
  year={2025}
}

@article{vflow,
  title={Vflow: Discovering optimal agentic workflows for verilog generation},
  author={Wei, Yangbo and Huang, Zhen and Li, Huang and Xing, Wei W and Lin, Ting-Jung and He, Lei},
  journal={arXiv preprint arXiv:2504.03723},
  year={2025}
}

@article{verireason,
  title={VeriReason: Reinforcement Learning with Testbench Feedback for Reasoning-Enhanced Verilog Generation},
  author={Wang, Yiting and Sun, Guoheng and Ye, Wanghao and Qu, Gang and Li, Ang},
  journal={arXiv preprint arXiv:2505.11849},
  year={2025}
}

@article{revolution,
  title={REvolution: An Evolutionary Framework for RTL Generation driven by Large Language Models},
  author={Min, Kyungjun and Cho, Kyumin and Jang, Junhwan and Kang, Seokhyeong},
  journal={arXiv preprint arXiv:2510.21407},
  year={2025}
}

@article{chainofthought,
  title={Chain-of-thought prompting elicits reasoning in large language models},
  author={Wei, Jason and Wang, Xuezhi and Schuurmans, Dale and Bosma, Maarten and Xia, Fei and Chi, Ed and Le, Quoc V and Zhou, Denny and others},
  journal={Advances in neural information processing systems},
  volume={35},
  pages={24824--24837},
  year={2022}
}

@article{deepseekr1,
  title={Deepseek-r1: Incentivizing reasoning capability in llms via reinforcement learning},
  author={Guo, Daya and Yang, Dejian and Zhang, Haowei and Song, Junxiao and Zhang, Ruoyu and Xu, Runxin and Zhu, Qihao and Ma, Shirong and Wang, Peiyi and Bi, Xiao and others},
  journal={arXiv preprint arXiv:2501.12948},
  year={2025}
}

@article{gptoss,
  title={gpt-oss-120b \& gpt-oss-20b model card},
  author={Agarwal, Sandhini and Ahmad, Lama and Ai, Jason and Altman, Sam and Applebaum, Andy and Arbus, Edwin and Arora, Rahul K and Bai, Yu and Baker, Bowen and Bao, Haiming and others},
  journal={arXiv preprint arXiv:2508.10925},
  year={2025}
}

@inproceedings{rtllm,
  title={Openllm-rtl: Open dataset and benchmark for llm-aided design rtl generation},
  author={Liu, Shang and Lu, Yao and Fang, Wenji and Li, Mengming and Xie, Zhiyao},
  booktitle={Proceedings of the 43rd IEEE/ACM International Conference on Computer-Aided Design},
  pages={1--9},
  year={2024}
}

@inproceedings{dolphin,
  title={Dolphin: moving towards closed-loop auto-research through thinking, practice, and feedback},
  author={Yuan, Jiakang and Yan, Xiangchao and Zhang, Bo and Chen, Tao and Shi, Botian and Ouyang, Wanli and Qiao, Yu and Bai, Lei and Zhou, Bowen},
  booktitle={Proceedings of the 63rd Annual Meeting of the Association for Computational Linguistics (Volume 1: Long Papers)},
  pages={21768--21789},
  year={2025}
}

@inproceedings{pyranet,
  title={Pyranet: A multi-layered hierarchical dataset for verilog},
  author={Nadimi, Bardia and Boutaib, Ghali Omar and Zheng, Hao},
  booktitle={2025 62nd ACM/IEEE Design Automation Conference (DAC)},
  pages={1--7},
  year={2025},
  organization={IEEE}
}

@inproceedings{rtlrewriter,
  title={Rtlrewriter: Methodologies for large models aided rtl code optimization},
  author={Yao, Xufeng and Wang, Yiwen and Li, Xing and Lian, Yingzhao and Chen, Ran and Chen, Lei and Yuan, Mingxuan and Xu, Hong and Yu, Bei},
  booktitle={Proceedings of the 43rd IEEE/ACM International Conference on Computer-Aided Design},
  pages={1--7},
  year={2024}
}

@inproceedings{yosys,
  title={Yosys-a free Verilog synthesis suite},
  author={Wolf, Clifford and Glaser, Johann and Kepler, Johannes},
  booktitle={Proceedings of the 21st Austrian Workshop on Microelectronics (Austrochip)},
  volume={97},
  year={2013}
}

@inproceedings{MCTS,
  author            = {Coulom, R{\'e}mi},
  title             = {Efficient Selectivity and Backup Operators in Monte-Carlo Tree Search},
  booktitle         = {Computers and Games},
  year              = {2007},
  comment-publisher = {Springer Berlin Heidelberg},
  comment-address   = {Berlin, Heidelberg},
  pages             = {72--83},
  isbn              = {978-3-540-75538-8},
  doi               = {10.1007/978-3-540-75538-8_7}
}

@inproceedings{UCT,
  author            = {Kocsis, Levente and Szepesv{\'a}ri, Csaba},
  title             = {Bandit Based Monte-Carlo Planning},
  booktitle         = {Machine Learning: ECML 2006},
  year              = {2006},
  comment-publisher = {Springer Berlin Heidelberg},
  comment-address   = {Berlin, Heidelberg},
  pages             = {282--293},
  isbn              = {978-3-540-46056-5},
  doi               = {10.1007/11871842_29}
}

@misc{mgverilog,
      title={MG-Verilog: Multi-grained Dataset Towards Enhanced LLM-assisted Verilog Generation}, 
      author={Yongan Zhang and Zhongzhi Yu and Yonggan Fu and Cheng Wan and Yingyan Celine Lin},
      year={2024},
      eprint={2407.01910},
      archivePrefix={arXiv},
      primaryClass={cs.LG},
      url={https://arxiv.org/abs/2407.01910}, 
}

@misc{mage,
  title={Mage: A multi-agent engine for automated rtl code generation. In 2025 62nd ACM/IEEE Design Automation Conference (DAC)},
  author={Zhao, Yujie and Zhang, Hejia and Huang, Hanxian and Yu, Zhongming and Zhao, Jishen},
  year={2025},
  publisher={IEEE}
}

@misc{aider,
    author = {Gauthier, Paul},
    title = {Aider: AI pair programming in your terminal},
	copyright = {Apache-2.0},
	url = {https://github.com/Aider-AI/aider},
	publisher = {Aider AI},
	year = {2025},
    version = {0.86.0}
}

@misc{metagpt,
      title={MetaGPT: Meta Programming for A Multi-Agent Collaborative Framework}, 
      author={Sirui Hong and Mingchen Zhuge and Jiaqi Chen and Xiawu Zheng and Yuheng Cheng and Ceyao Zhang and Jinlin Wang and Zili Wang and Steven Ka Shing Yau and Zijuan Lin and Liyang Zhou and Chenyu Ran and Lingfeng Xiao and Chenglin Wu and Jürgen Schmidhuber},
      year={2024},
      eprint={2308.00352},
      archivePrefix={arXiv},
      primaryClass={cs.AI},
      url={https://arxiv.org/abs/2308.00352}, 
}

@misc{agilecoder,
      title={AgileCoder: Dynamic Collaborative Agents for Software Development based on Agile Methodology}, 
      author={Minh Huynh Nguyen and Thang Phan Chau and Phong X. Nguyen and Nghi D. Q. Bui},
      year={2024},
      eprint={2406.11912},
      archivePrefix={arXiv},
      primaryClass={cs.SE},
      url={https://arxiv.org/abs/2406.11912}, 
}

@misc{grpo,
      title={DeepSeekMath: Pushing the Limits of Mathematical Reasoning in Open Language Models}, 
      author={Zhihong Shao and Peiyi Wang and Qihao Zhu and Runxin Xu and Junxiao Song and Xiao Bi and Haowei Zhang and Mingchuan Zhang and Y. K. Li and Y. Wu and Daya Guo},
      year={2024},
      eprint={2402.03300},
      archivePrefix={arXiv},
      primaryClass={cs.CL},
      url={https://arxiv.org/abs/2402.03300}, 
}

@misc{verilogcoder,
      title={VerilogCoder: Autonomous Verilog Coding Agents with Graph-based Planning and Abstract Syntax Tree (AST)-based Waveform Tracing Tool}, 
      author={Chia-Tung Ho and Haoxing Ren and Brucek Khailany},
      year={2025},
      eprint={2408.08927},
      archivePrefix={arXiv},
      primaryClass={cs.AI},
      url={https://arxiv.org/abs/2408.08927}, 
}

@misc{llmverilogsurvey,
      title={Large Language Model for Verilog Code Generation: Literature Review and the Road Ahead}, 
      author={Guang Yang and Wei Zheng and Xiang Chen and Dong Liang and Peng Hu and Yukui Yang and Shaohang Peng and Zhenghan Li and Jiahui Feng and Xiao Wei and Kexin Sun and Deyuan Ma and Haotian Cheng and Yiheng Shen and Xing Hu and Terry Yue Zhuo and David Lo},
      year={2025},
      eprint={2512.00020},
      archivePrefix={arXiv},
      primaryClass={cs.AR},
      url={https://arxiv.org/abs/2512.00020}, 
}

@misc{TaiwanICContest,
  author       = {{Ministry of Education, Taiwan}},
  title        = {National University Integrated Circuit Design Contest},
  howpublished = {\url{https://proj.moe.edu.tw/moeisoc/cl.aspx?n=6147}},
  year         = {2025},
  note         = {Accessed: 2026-01-19}
}

@misc{ace,
      title={Agentic Context Engineering: Evolving Contexts for Self-Improving Language Models}, 
      author={Qizheng Zhang and Changran Hu and Shubhangi Upasani and Boyuan Ma and Fenglu Hong and Vamsidhar Kamanuru and Jay Rainton and Chen Wu and Mengmeng Ji and Hanchen Li and Urmish Thakker and James Zou and Kunle Olukotun},
      year={2025},
      eprint={2510.04618},
      archivePrefix={arXiv},
      primaryClass={cs.LG},
      url={https://arxiv.org/abs/2510.04618}, 
}
\newpage
\appendix

\section{Detailed Evolutionary Framework} \label{sec:general_framework}
The evolutionary process, formalized in Algorithm~\ref{alg:general_framework}, unifies the search strategies (IGR and MCTS) under a single optimization loop. As defined in the methodology, the search space consists of a set of nodes $\mathcal{N}$, where each node $N = (V, S, F)$ contains the Verilog code, the quantitative score derived from the STG, and the diagnostic feedback.

\begin{algorithm}[H]
    \caption{General Evolutionary Framework for EvolVE}
    \label{alg:general_framework}
    \begin{algorithmic}[1]
        \STATE \textbf{Input:} Problem description $D$, Testbench $T$
        \STATE \textbf{Output:} Optimized node $N^*$
        \STATE \textbf{Parameters:} Evaluation function $E(V,T)$, LLM $M$, Max Nodes $L_{max}$
        \STATE
        \STATE Initialize node set $\mathcal{N}$ with seed code generated from $D$.
        \STATE Initialize archive of best solutions $\mathcal{A} \leftarrow \emptyset$.
        \STATE $n_{count} \leftarrow |\mathcal{N}|$
        \STATE
        \WHILE{($S_{max} < 1.0$ \textbf{or} Task is Opt) \textbf{and} ($n_{count} < L_{max}$)}
            \STATE \COMMENT{Step 1: Parent Selection via Strategy (IGR or MCTS)}
            \STATE Select parent node $N_{\text{parent}} = (V_p, S_p, F_p)$ from $\mathcal{N}$.
            \STATE
            \STATE \COMMENT{Step 2: Child Generation}
            \STATE Construct $prompt$ using $D, V_p,$ score $S_p$ and feedback $F_p$.
            \STATE Generate child code $V_{\text{child}} \leftarrow M(prompt)$.
            \STATE
            \STATE \COMMENT{Step 3: Evaluation via STG}
            \STATE Compute score and feedback: $(S_{\text{child}}, F_{\text{child}}) \leftarrow E(V_{\text{child}}, T)$.
            \STATE Create new node $N_{\text{child}} = (V_{\text{child}}, S_{\text{child}}, F_{\text{child}})$.
            \STATE
            \STATE \COMMENT{Step 4: Update State and Archive}
            \STATE Update $\mathcal{N} \leftarrow \mathcal{N} \cup \{N_{\text{child}}\}$.
            \STATE Update $\mathcal{A}$ with $N_{\text{child}}$ if $S_{\text{child}}$ improves.
            \STATE $n_{count} \leftarrow n_{count} + 1$
        \ENDWHILE
        \STATE \textbf{return} The node $N^*$ from $\mathcal{A}$ with the highest score.
    \end{algorithmic}
\end{algorithm}

\section{Benchmark Improvements} \label{sec:modified-verilogeval2}
\paragraph{Mod-VerilogEval v2.}
VerilogEval v2 is a widely adopted benchmark for evaluating Verilog generation models, comprising 156 problems designed to assess hardware design capabilities. For our evaluation, we utilize the \texttt{spec-to-rtl} subset. 

However, our preliminary analysis revealed critical flaws in the original dataset, including inaccurate problem descriptions, unsynthesizable syntax, and logical inconsistencies in the reference models. These issues frequently led to correct designs being penalized, hindering effective evolutionary optimization. To address this, we collaborated with experienced IC engineers to create Mod-VerilogEval v2. This revision ensures every problem is clearly defined, syntactically compliant with standard Verilog, and logically solvable, thereby providing a stable ground truth for the evolutionary search.
Our specific improvements include:

\begin{itemize} 
    \item \textbf{Initialization and Reset Handling:} We removed non-synthesizable \texttt{initial} blocks from all reference designs. To ensure proper register initialization, we introduced an explicit \texttt{reset} signal and updated the corresponding problem prompts (e.g., Problems 34, 53, 66, 104). 
    \item \textbf{Synthesizability and Logic Corrections:} In several problems (e.g., 116, 124, 151, 156), the reference designs produced unsynthesizable high-impedance or undefined outputs. We rewrote these modules to adhere to standard synthesizable SystemVerilog conventions.
    \item \textbf{Specification Alignment and Disambiguation:} We resolved inconsistencies between prompts and implementations. For Problem 62, the module logic was corrected to match the selection requirement (selecting \texttt{a} or \texttt{b} when \texttt{sel} is 1). For Problem 63, we explicitly defined the shift register's direction in the prompt to eliminate ambiguity. 
    \item \textbf{FSM and Interface Refinement:} For problems involving Finite State Machines (FSMs) and complex interfaces (e.g., Problems 89, 93, 123, 134), we refined the descriptions to clarify state transition conditions and port declarations, removing extraneous information. 
    \item \textbf{Naming Standardization:} We corrected naming inconsistencies to ensure uniform testing. This includes fixing port mappings in Problem 99 (correcting \texttt{Y2}/\texttt{Y4} to \texttt{Y3}/\texttt{Y1}), resolving top-module naming issues in Problem 130.
\end{itemize}

\section{Siliconmind 7b Model} \label{sec:siliconmind-7b}
Siliconmind-7B is the lightweight, open-source 7-billion parameter student model forming the core of an end-to-end LLM system designed to generate functionally correct and PPA optimized Verilog modules from natural language. The system is developed to overcome challenges in the semiconductor industry, such as the scarcity of high-quality Verilog data and the need to train and serve on limited, on-site computational resources. Its training begins with data augmentation: a larger teacher model, DeepSeek-R1, filters and expands public datasets into high-quality samples that include $<problem, reasoning, code, testbench>$ components. Siliconmind-7B is then enhanced through Supervised Fine-Tuning (SFT) and GRPO using this augmented data to learn advanced Verilog design capabilities.

\section{IC-RTL Detailed Results} \label{sec:ic_rtl_detailed}

This section presents the detailed performance metrics for the IC-RTL benchmark. The tables below are arranged in parallel to maximize space efficiency.

\begin{table}[H]
    \scriptsize
    \setlength{\tabcolsep}{2pt}
    \centering
    
    \begin{minipage}[t]{0.49\textwidth}
        \centering
        \caption{Q1\_LBP Results}
        \label{tab:q1_lbp_results}
        \begin{tabular}{cccccc}
            \toprule
            \textbf{Cyc} & \textbf{Area} & \textbf{Time} & \textbf{Pwr} & \textbf{PPA} & \textbf{AT} \\ 
            ($ns$) & ($\mu m^2$) & ($ns$) & (mW) & & \\
            \midrule
            \multicolumn{6}{c}{\textbf{Initial Implementation}} \\
            \midrule
            3 & 6.49E+03 & 1.93E+05 & 0.29 & 3.68E+08 & 4.19E+08 \\
            4 & 6.06E+03 & 2.58E+05 & 0.29 & 4.67E+08 & 3.91E+08 \\
            5 & 6.05E+03 & 3.22E+05 & 0.29 & 5.75E+08 & 3.90E+08 \\
            6 & 6.05E+03 & 3.87E+05 & 0.29 & 6.90E+08 & 3.90E+08 \\
            7 & 6.05E+03 & 4.51E+05 & 0.29 & 8.05E+08 & 3.90E+08 \\
            \midrule
            \multicolumn{6}{c}{\textbf{Optimized Version}} \\
            \midrule
            3 & 6.56E+03 & 1.93E+05 & 0.29 & 3.75E+08 & 4.23E+08 \\
            4 & 6.06E+03 & 2.58E+05 & 0.29 & 4.59E+08 & 3.91E+08 \\
            5 & 6.05E+03 & 3.22E+05 & 0.29 & 5.65E+08 & 3.90E+08 \\
            6 & 6.05E+03 & 3.87E+05 & 0.29 & 6.79E+08 & 3.90E+08 \\
            7 & 6.05E+03 & 4.51E+05 & 0.29 & 7.92E+08 & 3.90E+08 \\
            \bottomrule
        \end{tabular}
    \end{minipage}%
    \hfill
    \begin{minipage}[t]{0.49\textwidth}
        \centering
        \caption{Q2\_GEMM Results}
        \label{tab:q2_gemm_results}
        \begin{tabular}{cccccc}
            \toprule
            \textbf{Cyc} & \textbf{Area} & \textbf{Time} & \textbf{Pwr} & \textbf{PPA} & \textbf{AT} \\ 
            ($ns$) & ($\mu m^2$) & ($ns$) & (mW) & & \\
            \midrule
            \multicolumn{6}{c}{\textbf{Initial Implementation}} \\
            \midrule
            3 & 3.12E+05 & 1.08E+03 & 0.66 & 2.25E+08 & 1.13E+08 \\
            4 & 3.39E+05 & 1.45E+03 & 0.68 & 3.36E+08 & 1.23E+08 \\
            5 & 2.90E+05 & 1.81E+03 & 0.55 & 2.91E+08 & 1.05E+08 \\
            6 & 2.47E+05 & 2.17E+03 & 0.50 & 2.73E+08 & 8.98E+07 \\
            7 & 2.34E+05 & 2.54E+03 & 0.50 & 3.02E+08 & 8.52E+07 \\
            \midrule
            \multicolumn{6}{c}{\textbf{Optimized Version}} \\
            \midrule
            3 & 3.14E+05 & 9.63E+02 & 0.64 & 1.96E+08 & 1.01E+08 \\
            4 & 3.15E+05 & 1.28E+03 & 0.67 & 2.73E+08 & 1.01E+08 \\
            5 & 2.96E+05 & 1.60E+03 & 0.54 & 2.57E+08 & 9.50E+07 \\
            6 & 2.49E+05 & 1.92E+03 & 0.50 & 2.44E+08 & 8.01E+07 \\
            7 & 2.35E+05 & 2.24E+03 & 0.50 & 2.69E+08 & 7.56E+07 \\
            \bottomrule
        \end{tabular}
    \end{minipage}
\end{table}

\begin{table}[H]
    \scriptsize
    \setlength{\tabcolsep}{2pt}
    \centering
    
    \begin{minipage}[t]{0.49\textwidth}
        \centering
        \caption{Q3\_CONV Results}
        \label{tab:q3_conv_results}
        \begin{tabular}{cccccc}
            \toprule
            \textbf{Cyc} & \textbf{Area} & \textbf{Time} & \textbf{Pwr} & \textbf{PPA} & \textbf{AT} \\ 
            ($ns$) & ($\mu m^2$) & ($ns$) & (mW) & & \\
            \midrule
            \multicolumn{6}{c}{\textbf{Initial Implementation}} \\
            \midrule
            3 & 3.64E+04 & 1.78E+05 & 0.59 & 3.83E+08 & 6.49E+09 \\
            4 & 3.63E+04 & 2.37E+05 & 0.57 & 4.93E+08 & 8.51E+09 \\
            5 & 3.00E+04 & 2.97E+05 & 0.51 & 4.58E+08 & 8.93E+09 \\
            6 & 2.35E+04 & 3.56E+05 & 0.47 & 3.93E+08 & 8.37E+09 \\
            7 & 2.36E+04 & 4.16E+05 & 0.46 & 4.52E+08 & 9.84E+09 \\
            \midrule
            \multicolumn{6}{c}{\textbf{Optimized Version}} \\
            \midrule
            3 & 3.58E+04 & 1.78E+05 & 0.69 & 3.92E+08 & 6.37E+09 \\
            4 & 3.53E+04 & 2.38E+05 & 0.62 & 5.15E+08 & 8.38E+09 \\
            5 & 3.09E+04 & 2.97E+05 & 0.55 & 5.02E+08 & 9.18E+09 \\
            6 & 2.33E+04 & 3.56E+05 & 0.50 & 4.18E+08 & 8.30E+09 \\
            7 & 2.23E+04 & 4.16E+05 & 0.50 & 4.67E+08 & 9.29E+09 \\
            \bottomrule
        \end{tabular}
    \end{minipage}%
    \hfill
    \begin{minipage}[t]{0.49\textwidth}
        \centering
        \caption{Q4\_JAM Results}
        \label{tab:q4_jam_results}
        \begin{tabular}{cccccc}
            \toprule
            \textbf{Cyc} & \textbf{Area} & \textbf{Time} & \textbf{Pwr} & \textbf{PPA} & \textbf{AT} \\ 
            ($ns$) & ($\mu m^2$) & ($ns$) & (mW) & & \\
            \midrule
            \multicolumn{6}{c}{\textbf{Initial Implementation}} \\
            \midrule
            3 & 1.31E+04 & 1.21E+06 & 0.82 & 1.30E+10 & 5.27E+09 \\
            4 & 1.09E+04 & 1.61E+06 & 0.77 & 1.35E+10 & 4.38E+09 \\
            5 & 9.14E+03 & 2.02E+06 & 0.53 & 9.83E+09 & 3.69E+09 \\
            6 & 8.98E+03 & 2.42E+06 & 0.44 & 9.59E+09 & 3.62E+09 \\
            7 & 8.97E+03 & 2.82E+06 & 0.37 & 9.34E+09 & 3.62E+09 \\
            \midrule
            \multicolumn{6}{c}{\textbf{Optimized Version}} \\
            \midrule
            3 & 1.30E+04 & 1.21E+06 & 1.05 & 1.66E+10 & 5.26E+09 \\
            4 & 1.01E+04 & 1.61E+06 & 0.61 & 9.91E+09 & 4.07E+09 \\
            5 & 8.77E+03 & 2.02E+06 & 0.41 & 7.32E+09 & 3.54E+09 \\
            6 & 8.74E+03 & 2.42E+06 & 0.34 & 7.22E+09 & 3.53E+09 \\
            7 & 8.73E+03 & 2.82E+06 & 0.29 & 7.18E+09 & 3.52E+09 \\
            \bottomrule
        \end{tabular}
    \end{minipage}
\end{table}

\begin{table}[H]
    \scriptsize
    \setlength{\tabcolsep}{2pt}
    \centering
    
    \begin{minipage}[t]{0.49\textwidth}
        \centering
        \caption{Q5\_HC Results}
        \label{tab:q5_hc_results}
        \begin{tabular}{cccccc}
            \toprule
            \textbf{Cyc} & \textbf{Area} & \textbf{Time} & \textbf{Pwr} & \textbf{PPA} & \textbf{AT} \\ 
            ($ns$) & ($\mu m^2$) & ($ns$) & (mW) & & \\
            \midrule
            \multicolumn{6}{c}{\textbf{Initial}} \\
            \midrule
            3 & 2.39E+04 & 4.29E+02 & 2.49 & 2.56E+07 & 3.43E+06 \\
            4 & 1.98E+04 & 5.72E+02 & 1.41 & 1.60E+07 & 2.84E+06 \\
            5 & 1.84E+04 & 7.15E+02 & 1.05 & 1.38E+07 & 2.63E+06 \\
            6 & 1.78E+04 & 8.58E+02 & 0.85 & 1.30E+07 & 2.55E+06 \\
            7 & 1.80E+04 & 1.00E+03 & 0.73 & 1.32E+07 & 2.58E+06 \\
            \midrule
            \multicolumn{6}{c}{\textbf{Area Opt.}} \\
            \midrule
            3 & 1.82E+04 & 4.29E+02 & 1.97 & 1.54E+07 & 2.61E+06 \\
            4 & 1.54E+04 & 5.72E+02 & 1.17 & 1.03E+07 & 2.20E+06 \\
            5 & 1.48E+04 & 7.15E+02 & 0.90 & 9.56E+06 & 2.11E+06 \\
            6 & 1.45E+04 & 8.58E+02 & 0.71 & 8.94E+06 & 2.07E+06 \\
            7 & 1.44E+04 & 1.00E+03 & 0.61 & 8.92E+06 & 2.06E+06 \\
            \midrule
            \multicolumn{6}{c}{\textbf{Cycle Opt.}} \\
            \midrule
            3 & 2.92E+04 & 3.39E+02 & 2.25 & 2.23E+07 & 3.30E+06 \\
            4 & 2.80E+04 & 4.52E+02 & 2.42 & 3.07E+07 & 3.17E+06 \\
            5 & 2.74E+04 & --- & --- & --- & --- \\
            6 & 2.79E+04 & 6.78E+02 & 1.19 & 2.26E+07 & 3.15E+06 \\
            7 & 2.64E+04 & 7.91E+02 & 1.61 & 3.37E+07 & 2.99E+06 \\
            \bottomrule
        \end{tabular}
    \end{minipage}%
    \hfill
    \begin{minipage}[t]{0.49\textwidth}
        \centering
        \caption{Q6\_DT Results}
        \label{tab:q6_dt_results}
        \begin{tabular}{cccccc}
            \toprule
            \textbf{Cyc} & \textbf{Area} & \textbf{Time} & \textbf{Pwr} & \textbf{PPA} & \textbf{AT} \\ 
            ($ns$) & ($\mu m^2$) & ($ns$) & (mW) & & \\
            \midrule
            \multicolumn{6}{c}{\textbf{Initial Implementation}} \\
            \midrule
            3 & 1.10E+04 & 3.59E+05 & 0.53 & 2.14E+09 & 1.32E+09 \\
            4 & 9.12E+03 & 4.78E+05 & 0.37 & 1.63E+09 & 1.09E+09 \\
            5 & 8.01E+03 & 5.98E+05 & 0.34 & 1.64E+09 & 9.59E+09 \\
            6 & 8.19E+03 & 7.18E+05 & 0.36 & 2.13E+09 & 9.80E+08 \\
            7 & 7.89E+03 & 8.38E+05 & 0.34 & 2.26E+09 & 9.45E+08 \\
            \midrule
            \multicolumn{6}{c}{\textbf{Area Optimized}} \\
            \midrule
            3 & 1.01E+04 & 3.59E+05 & 0.52 & 1.89E+09 & 1.21E+09 \\
            4 & 8.82E+03 & 4.78E+05 & 0.40 & 1.70E+09 & 1.05E+09 \\
            5 & 7.60E+03 & 5.98E+05 & 0.34 & 1.57E+09 & 9.11E+08 \\
            6 & 7.81E+03 & 7.18E+05 & 0.36 & 2.06E+09 & 9.36E+08 \\
            7 & 7.47E+03 & 8.38E+05 & 0.34 & 2.18E+09 & 8.95E+08 \\
            \midrule
            \multicolumn{6}{c}{\textbf{Cycle Optimized}} \\
            \midrule
            3 & 1.07E+04 & 3.48E+05 & 0.52 & 1.94E+09 & 1.24E+09 \\
            4 & 9.36E+03 & 4.65E+05 & 0.38 & 1.67E+09 & 1.08E+09 \\
            5 & 8.12E+03 & 5.81E+05 & 0.35 & 1.66E+09 & 9.44E+08 \\
            6 & 8.28E+03 & 6.97E+05 & 0.36 & 2.09E+09 & 9.63E+08 \\
            7 & 7.95E+03 & 8.13E+05 & 0.34 & 2.22E+09 & 9.25E+08 \\
            \bottomrule
        \end{tabular}
    \end{minipage}
\end{table}

\end{document}